\documentclass[]{article}
\usepackage[affil-it]{authblk}
\usepackage{fullpage}
\usepackage{soul}
\usepackage{url}

\usepackage[utf8]{inputenc}
\usepackage[small]{caption}
\usepackage{graphicx}
\usepackage{amsmath,amsthm,amssymb}
\usepackage{booktabs}
\usepackage{hyperref}

\newtheorem{theorem}{Theorem}
\newtheorem{definition}[theorem]{Definition}
\usepackage[numbers]{natbib} 
\usepackage{xspace}

\newcommand{\loc}{J}
\newcommand{\cus}{I}

\newcommand{\PCP}{pCP\xspace}

\usepackage{xcolor}

\begin{document}

\title{Experiments with graph convolutional networks for solving the vertex $p$-center problem}

\author[1]{Elisabeth Gaar\thanks{elisabeth.gaar@jku.at}}

\affil[1]{Institute of Production and Logistics Management, Johannes Kepler University Linz, Linz, Austria}

\author[1,2]{Markus Sinnl\thanks{markus.sinnl@jku.at}}

\affil[2]{JKU Business School, Johannes Kepler University Linz, Linz, Austria}

\date{}

\maketitle

\begin{abstract}
	
In the last few years, graph convolutional networks (GCN) have become a popular 
research direction in the machine learning community to tackle NP-hard 
combinatorial optimization problems (COPs) defined on graphs. While the 
obtained results are usually still not competitive with problem-specific 
solution approaches from the operations research community, GCNs often lead to 
improvements compared to previous machine learning approaches for classical 
COPs such as the traveling salesperson problem (TSP). 

In this work we present a preliminary study on using GCNs for solving the 
vertex $p$-center problem (\PCP), which is another classic COP on graphs. In 
particular, we investigate whether a successful model based on 
\emph{end-to-end} 
training for the TSP can be adapted to a \PCP, which is defined on a 
similar 2D Euclidean graph input as the usually used version of the TSP. 
However, the 
objective of the \PCP has a min-max structure which could lead to many 
symmetric optimal, i.e., ground-truth solutions and other potential 
difficulties for learning. Our obtained preliminary results show that indeed a 
direct transfer of network architecture ideas does not seem to work too well. 
Thus we think that the \PCP could be an interesting benchmark problem for new 
ideas and developments in the area of GCNs.
\end{abstract}

\section{Introduction}

In the last few years, graph convolutional networks (GCN) have become a popular 
research direction in the machine learning community to tackle NP-hard 
combinatorial optimization problems (COPs) defined on graphs (see, e.g., 
\citep{cappart2021combinatorial,li2018combinatorial}). In the operations 
research (OR) community NP-hard problems are usually formulated as integer 
programming (IP) problems, however, due to the theoretical difficulty, even 
with sophisticated IP solution frameworks solving large instances of such 
problems may become computationally intractable. Moreover, such solution 
frameworks often also need carefully handcrafted heuristics and additional 
problem-specific knowledge to work well. Thus, alternative approaches such as 
machine learning algorithms which can be trained to directly learn the solution 
from problem instances could offer an attractive solution strategy 
\cite{bengio2020machine,smith1999neural}.

Using advances in graph representation learning including GCNs, several COPs, 
such as the \emph{travelling salesperson problem (TSP)} 
\cite{kool2018attention,joshi2019efficient,joshi2020learning}, the \emph{vertex 
cover problem}, the \emph{maximum cut problem} 
\cite{dai2017learning,li2018combinatorial} or \emph{graph matching} 
\cite{fey2020deep} have been approach by \emph{end-to-end-approaches} recently. 
However, the results in these works show that developing machine learning 
approaches that are competitive with specialized OR approaches is still an 
open question, see also the recent survey \cite{cappart2021combinatorial}.

In this work, we expand the existing literature on the use of GCN for COPs by 
applying it to the \emph{vertex-$p$-center problem (\PCP)}. The \PCP is a 
fundamental problem in location science (see, e.g., \cite{laporte2019location}) 
and is formally defined as follows.

\begin{definition}[Vertex $p$-center problem]
Let $p$ be an integer, $\cus$ be  a 
set of customer 
demand points with cardinality $|\cus|=n$ and $\loc$ be a set of potential 
facility 
locations  with cardinality $|\loc|=m \geq p$. Furthermore let $d$ be a 
distance 
function  such that $d_{ij}$ is the distance between each
	customer 
	demand point $i \in \cus $ and potential facility location $j \in \loc$. The goal is to
	find a subset $S \subseteq \loc$ of facilities with cardinality $|S|=p$ to 
	\emph{open} such that the maximum 
	distance between a customer demand point and its closest 
	open facility is minimized, i.e. such that 
	$\max_{i \in \cus} \min_{j \in S} \{d_{ij} \}$ is minimized.	
\end{definition}	

Following the standard convention used in the OR community, in the remainder of 
this paper we use $\cus=\loc=V$.
Similar to approaches for the TSP \cite{joshi2019efficient,joshi2020learning}, 
we focus on instances which are defined on the 2D Euclidean plane, i.e., each 
point $i \in V$ has coordinates $x_i\in \mathbb Z^2$. These coordinates will be 
used as features in our model, together with an underlying graph structure 
which is defined on the $k$-nearest neighbors for each vertex. The goal of our 
approach is to obtain a $p_i^S \in [0,1]$ for every potential facility 
location $i \in 
V$, 
such that $p_i^S$ represents the probability that $i$ is in 
the optimal solution.  These probabilities are then transformed into a 
feasible 
solution using \emph{post-hoc} techniques. We use \emph{supervised} training 
and 
the optimal solutions for our test instances are obtained by solving the 
classical IP formulation for the \PCP (see, e.g., \cite{laporte2019location}).
Our whole solution approach closely follows the approach proposed in 
\cite{joshi2019efficient,joshi2020learning} for the TSP, which is among the 
state-of-the-art machine learning approaches for the TSP. By doing so, we not 
only leverage existing work, but also try to investigate how well general 
machine learning architectures transfer between two COPs, where the input is 
given in a quite similar way. Aside from this, we believe there are several 
additional reasons why considering the \PCP is an attractive target to study in 
the machine learning context:
\begin{itemize}
	\item Due to the min-max structure of the objective function of the \PCP, 
	classical direct IP approaches which are usually used in the OR community 
	do not work well (see, e.g., \cite{snyder2011fundamentals}). Thus, 
	state-of-the-art exact solution approaches use the connection of the \PCP 
	to the \emph{set-cover problem}, and iteratively solve a series of 
	set-cover 
	problems, which are defined with a dependency on the objective function 
	value of the 
	\PCP. If a good objective 
	function value is known, these approaches can be speed up considerably, 
	and a successful machine learning approach could provide such a good 
	objective function value.
	\item Compared to the TSP, where optimal solutions to instances are usually 
	unique, there can exist many different optimal solutions for a \PCP 
	instance. This is again due to the min-max objective function. This could 
	make end-to-end training harder. Moreover there is also the given 
	cardinality constraint $p$ on the number of vertices in the solution, which 
	could in general make it harder for a machine learning algorithm to discern 
	which vertices should have a high probability to be in a solution, as there 
	will be ``symmetric'' vertices, where any of them could replace the other 
	without sacrificing solution quality.
	\item The cardinality constraint could also offer an attractive target for 
	\emph{transfer learning}, i.e, instead of just trying to transfer the 
	knowledge learned from a graph with few vertices to a graph with more 
	vertices, it could also be interesting to investigate if some transfer 
	between knowledge learned for different values of $p$ is possible.
	\item Finally, instead of trying to directly find a solution to the 
	problem, machine learning techniques could also be used for instance-size 
	reduction before an exact solution algorithm from the OR community is 
	applied. For example, in the state-of-the-art exact approach 
	\cite{contardo2019scalable} for the \PCP, it is enough to solve set-cover 
	problems defined on just a few hundred customers for instances with ten 
	thousands of vertices. In \cite{contardo2019scalable} these customers are 
	found using OR techniques (i.e., problem specific heuristics) during the 
	course of an algorithm. These heuristics could potentially be replaced 
	with machine learning approaches.
	
\end{itemize}

In this paper, we present preliminary work to partially address the points raised above.


\section{Model}

Our model consist of a graph convolutional network, and we output for each 
potential facility location $i\in V$ a probability $p^S_i$ of being in the 
optimal solution. This is achieved by a sigmoid output layer. In the 
test-phase, we use greedy-heuristics to convert the probabilities to valid 
solutions. In this section, we give a general description of the network. 
Details about the concrete number of layers and features used in our 
experiments are discussed in the next section.

\subsection{Input layer}

For each $i\in V$, we use the coordinates $x_i \in \mathbb Z^2$ as vertex-input 
features. They are transformed into $h$-dimensional features using a linear 
transformation, i.e.,
\begin{equation*}
x^0_{i}= A_1 x_i + b
\end{equation*}
where $A_1 \in \mathbb R^{h \times 2}$ and $b \in  \mathbb R^{h}$ are 
learnable parameters.

\subsection{Graph convolution layer}

As a graph convolution layer, we use the residual gated graph convolutional 
operator proposed by \cite{bresson2017residual}. When using such a graph 
convolutional operator, in order to represent 
the local graph structure for each vertex $i \in V$ the features of the 
neighbors of $i$ are gathered via recursive message passing. Aside from 
\cite{bresson2017residual}, similar operators 
where also presented in e.g., 
\cite{defferrard2016convolutional,li2015gated,kipf2016semi,hamilton2017inductive}.
 
For a given vertex $i \in V$, let $N(i)$ be the set of its neighbors in the 
input graph. Given the vertex features $x^\ell$ at layer~$\ell$, the vertex 
features $x^{\ell+1}$ at 
layer 
$\ell+1$ are defined as 
\begin{equation*}
x^{\ell+1}_{i}= x^\ell_i+ReLU(A^\ell_2 x^\ell_i+ \sum_{i' \in N(i)} 
\eta_{i,i'}^\ell \odot A_3^\ell x^\ell_{i'})
\end{equation*}
with
\begin{equation*}
\eta^\ell_{i,i'}=\sigma(A_4^\ell x_i^\ell +A_5^\ell x_{i'}^\ell), 
\end{equation*} 
where $ReLU$ denotes the rectified linear unit, $\sigma$ denotes the sigmoid 
function, $\odot$ denotes the Hadamard product, $A_2^\ell, A_3^\ell, A_4^\ell, 
A_5^\ell \in \mathbb 
R^{h \times h}$ are learnable parameters, and $\ell = 0, \dots, L-1$, i.e., 
$L$ is the index of the last convolution layer.

In our models, we redefine the 
neighborhood-structure $N(i)$ for a vertex $i \in V$ by including not all 
neighbors, but by including only the $k$-nearest neighbors with respect to the 
distance $d$ for a given integer $k$. We will report results obtained by using 
different 
values of $k$.

\subsection{Output layer}

To compute the probability $p^S_i$ of vertex $i \in V$ to be in the optimal 
solution, we 
use the vertex embedding $x^L_i$ of the last convolution layer with index $L$ 
and apply a 
multi-layer perceptron (MLP) followed by a sigmoid function, i.e., 
\begin{equation*}
p^S_i = \sigma(MLP(x^L_i)).
\end{equation*}

\subsection{Loss function}

We minimize the weighted cross-entropy loss between the ground-truth solution 
$\hat p_i$ (where $\hat p_i=1$ if and only if the location $i$ is in the 
solution an zero 
otherwise) and the obtained probabilities $p^S_i$. As the classification task 
is unbalanced (i.e., for each 
instance, there are $p$ vertices in the solution and $n-p$ outside), we give 
weight $(n-p)/p$ to the class of vertices $i$ such that $\hat p_i=1$.
 The loss is averaged over mini-batches.

\newcommand{\naive}{\texttt{naive}}
\newcommand{\greedy}{\texttt{greedy}}

\subsection{Solution decoding}

Given the output-probabilities $p^S_i$ for all $i \in V$ of our model, we 
implemented the following two strategies to convert the probabilities into a 
feasible solution $S^H$:

\begin{itemize}
	\item \naive: Sort the vertices according to probabilities in a descending 
	order (ties broken arbitrarily) and pick the $p$ vertices with the largest 
	probabilities to obtain a solution~$S^H$.
	\item \greedy: Due to the structure of the problem (i.e., the min-max 
	objective), several vertices can have a nearly identical effect on the 
	solution quality, and once one of these vertices is picked, it does not 
	pay-off to include any of the other ``similar'' vertices. To take this into 
	account, we tried the following greedy-strategy: Sort the vertices 
	according 
	to probabilities in a descending order (ties broken arbitrarily). Pick the 
	vertex with the largest probability and insert it into $S^H$. Then iterate 
	trough the remainder of the sorted list of vertices, but only add a vertex 
	to 
	$S^H$ if adding this vertex improves the objective function. Stop when 
	$|S^H|=p$.
\end{itemize}

\section{Experiments}

Our proposed model was implemented in \texttt{Python} using \texttt{PyTorch} \cite{paszke2019pytorch} and \texttt{PyTorch Geometric} \cite{fey2019fast}. The experiments were done on an AMD Ryzen 5 2600 with 3.4Ghz and six cores, and 16GB of RAM.

\subsection{Data set}

Similar to other work for COPs on graphs like the TSP (see, e.g., 
\cite{vinyals2015pointer,joshi2019efficient,joshi2020learning}), we train and 
evaluate our model on fixed-size problem instances. As mentioned above, we are 
following the standard in the OR community and let $\cus=\loc=V$, and we use 
instances with $n=m=50$. Moreover, we set $p=5$. We create each instance by 
randomly sampling $n$ points in the integer-square $[0,100]^2$. As distance 
$d_{ii'}$ between two vertices $i,i' \in V$, we use the Euclidean distance 
based on 
their coordinates. The training sets consists of 100,000 pairs of problem 
instances and optimal solutions, and the test set of 1,000 such pairs. The 
optimal solutions were calculated by solving the IP formulation using the IP 
solver \texttt{CPLEX} \cite{noauthor_cplex_nodate}. We note that the exact 
solution time for these instances is under one second on average.

\newcommand{\A}{\texttt{A}\xspace}
\newcommand{\B}{\texttt{B}\xspace}
\newcommand{\C}{\texttt{C}\xspace}

\subsection{Hyperparameter configuration}

We consider three different sets of hyperparameters in our experiments to 
investigate the influence of the neigborhood-size $k$ and the number of graph 
convolutional layers $L$. We note that by using $L$ graph convolutional layers, 
the network can use the $L$-hop neighborhood information.
We use the following three settings. 

\begin{itemize}
	\item setting \A: One graph convolutional layer, MLP with three layers, 
	$h=50$ (resulting in approximately 15,000 trainable parameters), 
	neighborhood-size $k=10$
	\item setting \B: Three graph convolutional layers, MLP with three layers, 
	$h=100$ (resulting in approximately 140,000 trainable parameters), 
	neighborhood-size $k=5$ 
	\item setting \C: Three graph convolutional layers, MLP with three layers, 
	$h=100$ (resulting in approximately 140,000 trainable parameters), 
	neighborhood-size $k~=~10$ 
\end{itemize}

\subsection{Training procedure}

\begin{figure}[h!]
	\centering
\includegraphics[width=0.5\columnwidth]{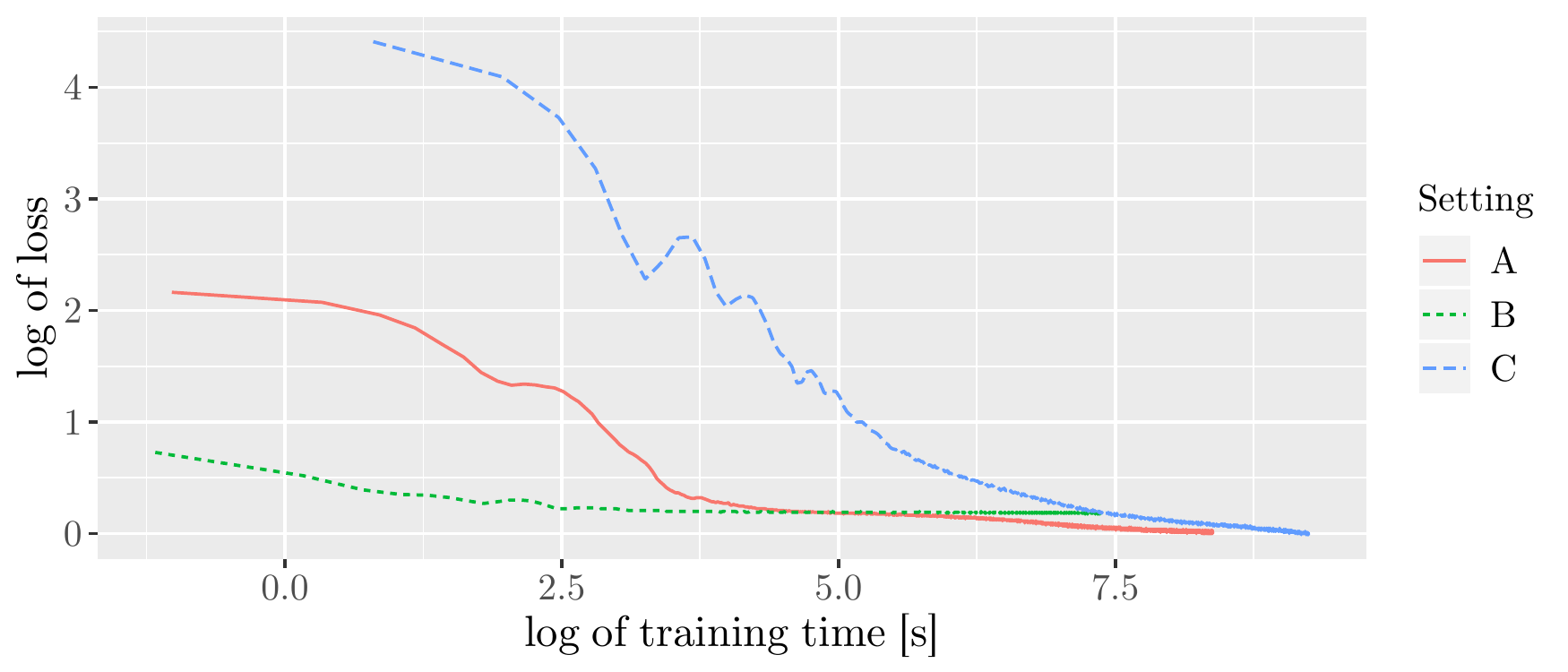}
\caption{Training-time against loss for the different settings.\label{fig:loss}}
\end{figure}

We train our model directly in an \emph{end-to-end} fashion by minimizing the 
loss function via gradient descent. We use the \texttt{Adam} optimizer 
\cite{kingma2014adam} with learning rate $0.0001$. The training is done using 
$100$ mini-batches of size $1,000$, setting \A is trained for 50 epochs, and \B 
and \C for 20 epochs (the latter have a larger number of trainable parameters 
and thus training takes longer). Figure \ref{fig:loss} shows a plot of the loss 
against training-time for all three settings. The training-time for setting \A 
was 4,361 seconds and the loss at the end of training was 1.01. For setting \B 
the training-time  
was 1,577 seconds and the loss at the end was 1.20. For setting \C the 
training-time was 
10,397 seconds and the loss at the end was 1.00. Thus it seems that larger $k$ 
leads to a lower loss, while the number of layers $L$ or features $h$ does not 
seem to 
be as crucial.

\subsection{Results and discussion}

\begin{figure}[h!]
	\centering
	\includegraphics[width=0.5\columnwidth]{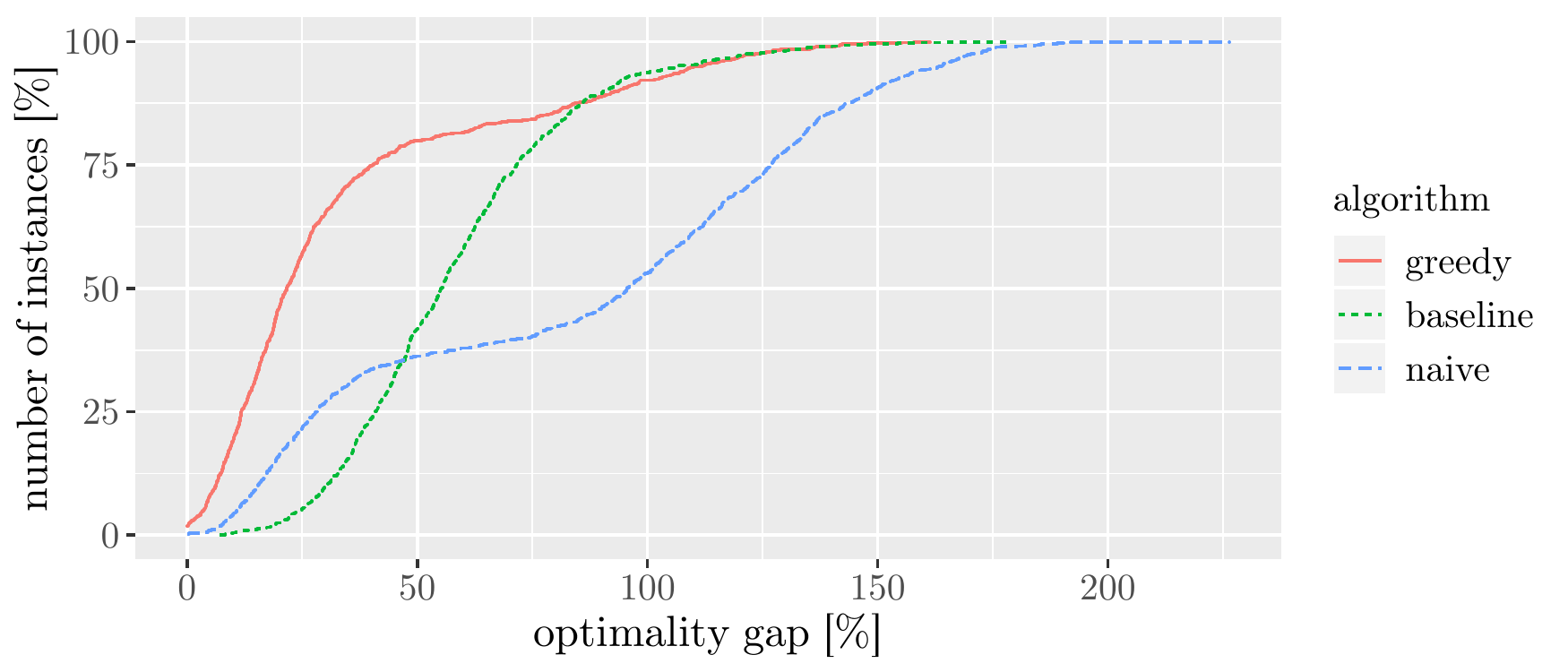}
	\caption{Plot of optimality gap for setting \A.\label{fig:perfA}}
\end{figure}

\begin{figure}[h!]
		\centering
	\includegraphics[width=0.5\columnwidth]{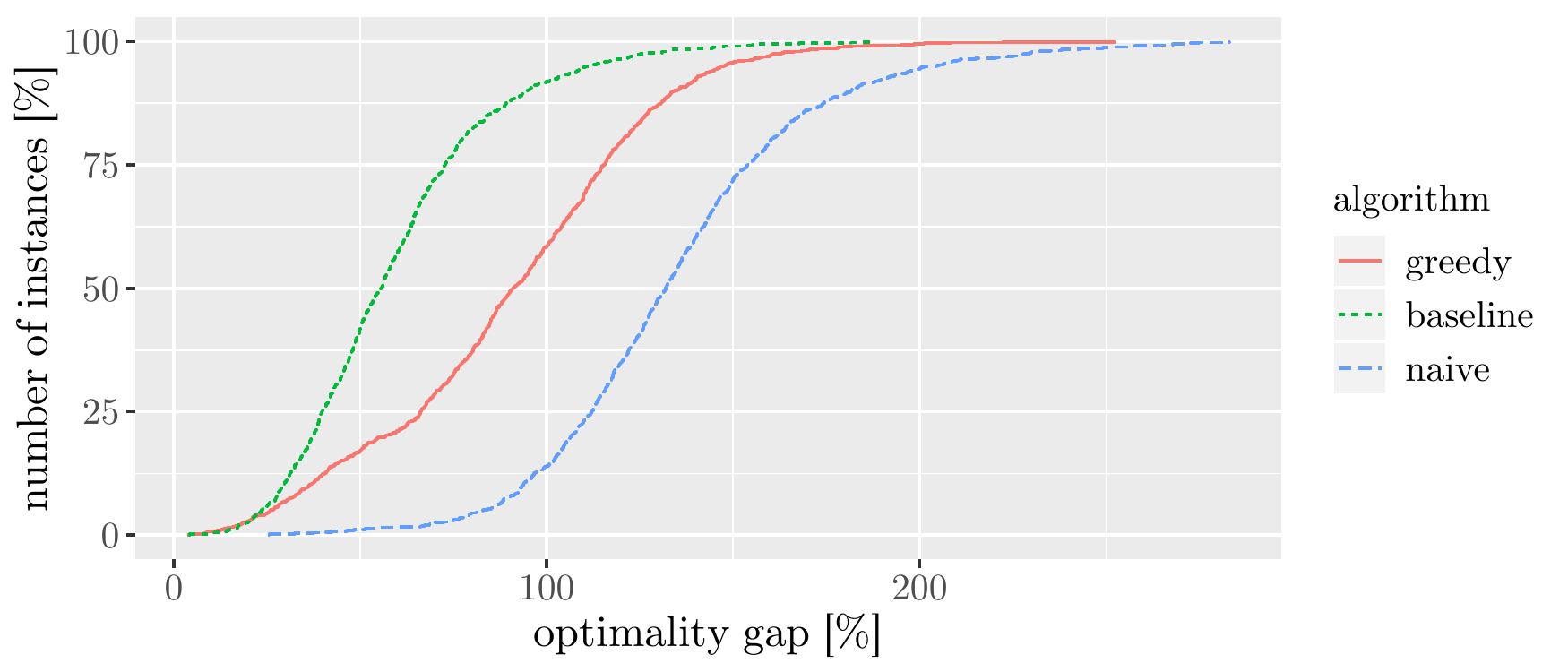}
	\caption{Plot of optimality gap for setting \B.\label{fig:perfB}}
\end{figure}

\begin{figure}[h!]
		\centering
	\includegraphics[width=0.5\columnwidth]{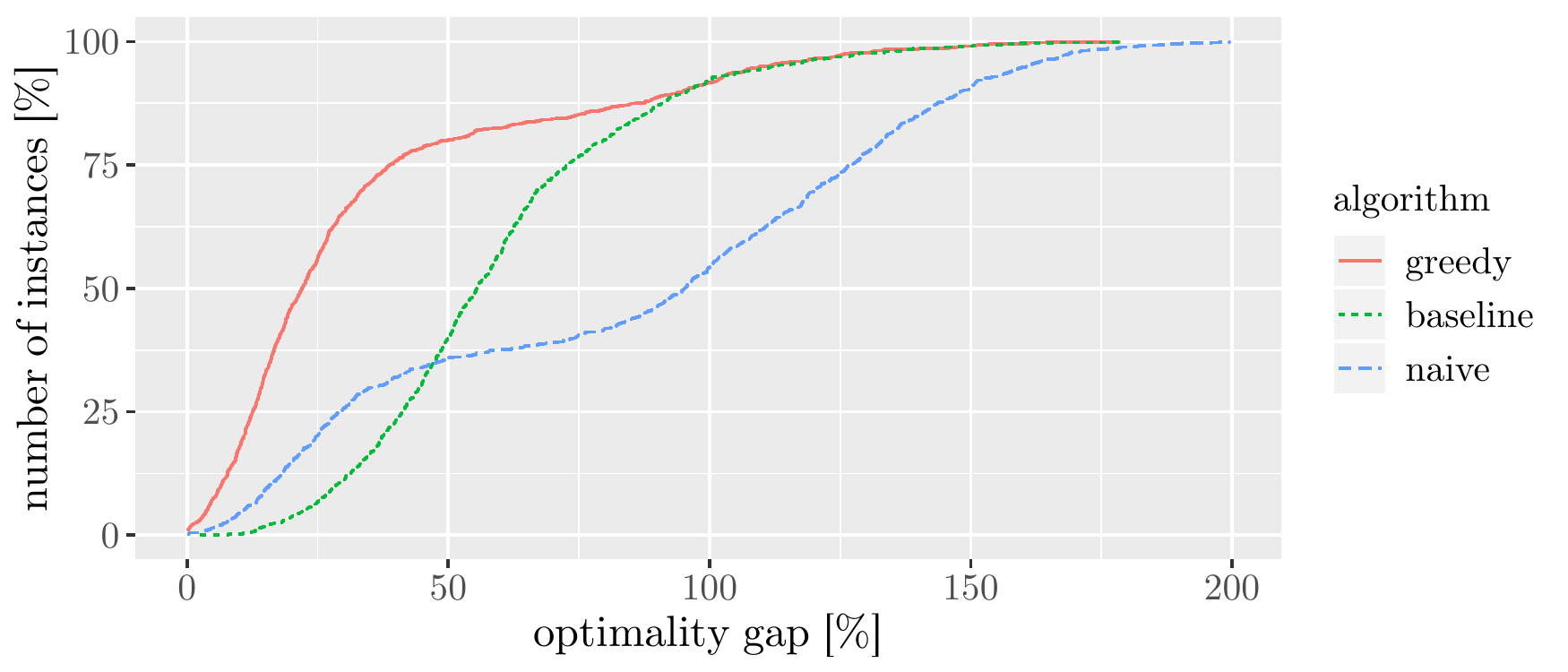}
	\caption{Plot of optimality gap for setting \C.\label{fig:perfC}}
\end{figure}

To investigate the effect of the information gained by our model, aside from 
comparing with the solution values $z^*$ of the ground-truth (i.e., optimal) 
solutions, we have also implemented a version of the greedy decoder algorithm 
\texttt{greedy}, which, instead of being initialized with the probabilities 
$p^S_i$ for all $i \in V$ obtained by our model, is initialized with random 
values as $p_i^S$. This baseline greedy algorithm is denoted by 
\texttt{baseline} 
and the obtained values are denoted by $z^B$. In 
Figures~\ref{fig:perfA},~\ref{fig:perfB}, and~\ref{fig:perfC} we present plots 
of the optimality gaps $(z-z^*)/z^*\cdot 100$ of the solution values $z=z^N, 
z^G, z^B$ of \texttt{naive}, \texttt{greedy}, \texttt{baseline}, respectively 
for the instances of our test-set and our different settings.

The results show that the \texttt{naive} decoding strategy works worse than 
even the random \texttt{baseline} algorithm for all three settings. This is not 
unexpected, because as 
discussed in the introduction also for OR-approaches, the ``symmetry'' of the 
optimal solutions of instances is often troublesome  and the \texttt{naive} 
decoding strategy does not take care of this issue. Thus, the \texttt{naive} 
decoding strategy could only work well if the obtained probabilities $p^S_i$ of 
our model would already encode some information about this. However, this seems 
not to be the case. We note that also for problems like the TSP, the 
combination of the obtained probabilities with a sophisticated post-hoc 
algorithm to obtain a feasible solution is crucial for a good performance, see 
e.g., \cite{joshi2019efficient,joshi2020learning}. The min-max structure of the 
\PCP may even increase this difficulty.
One potential path to overcome this issue of not being able to deal with 
multiple optimal solutions for the future is to not only compute one 
probability $p^S_i$ for all $i \in V$, but instead to compute multiple 
probabilities and then using a minimum operator when computing the loss 
function, like it is done in \cite{li2018combinatorial}.

Another interesting result which can be seen in the plot is that for settings 
\A and \C, which are both settings with $k~=~10$, the \texttt{greedy} approach 
leads to better solution quality for most of the instances compared to the 
\texttt{baseline}, while for setting \B, \texttt{baseline} outperforms 
\texttt{greedy}. This seems to be consistent with the value obtained for the 
loss function, i.e., using only $k=5$ leads to a larger loss-value at the end 
of 
the training compared to the two settings with $k=10$ and now in the evaluation 
the results show that in this case we did not seem to have learned anything 
useful.

In general, the optimality-gaps are rather large compared to results obtained 
e.g., for the TSP in \cite{joshi2019efficient,joshi2020learning}. Thus, it  
seems that directly using existing graph convolutional approaches for 
successfully solving the \PCP may not be possible. This could be due to the 
min-max objective function, which makes it different to many other existing 
COPs on graphs. We think the problem can be an interesting 
benchmark problem for the machine learning community as potentially new graph 
convolutional techniques or other advances need to be developed in order to 
obtain improved results. 

\bibliography{biblio}

\begin{thebibliography}{}

\bibitem[\protect\citeauthoryear{Bengio \bgroup \em et al.\egroup
  }{2020}]{bengio2020machine}
Yoshua Bengio, Andrea Lodi, and Antoine Prouvost.
\newblock Machine learning for combinatorial optimization: a methodological
  tour d’horizon.
\newblock {\em European Journal of Operational Research}, 2020.

\bibitem[\protect\citeauthoryear{Bresson and
  Laurent}{2017}]{bresson2017residual}
Xavier Bresson and Thomas Laurent.
\newblock Residual gated graph convnets.
\newblock {\em arXiv preprint arXiv:1711.07553}, 2017.

\bibitem[\protect\citeauthoryear{Cappart \bgroup \em et al.\egroup
  }{2021}]{cappart2021combinatorial}
Quentin Cappart, Didier Ch{\'e}telat, Elias Khalil, Andrea Lodi, Christopher
  Morris, and Petar Veli{\v{c}}kovi{\'c}.
\newblock Combinatorial optimization and reasoning with graph neural networks.
\newblock {\em arXiv preprint arXiv:2102.09544}, 2021.

\bibitem[\protect\citeauthoryear{Contardo \bgroup \em et al.\egroup
  }{2019}]{contardo2019scalable}
Claudio Contardo, Manuel Iori, and Raphael Kramer.
\newblock A scalable exact algorithm for the vertex p-center problem.
\newblock {\em Computers \& Operations Research}, 103:211--220, 2019.

\bibitem[\protect\citeauthoryear{Dai \bgroup \em et al.\egroup
  }{2017}]{dai2017learning}
Hanjun Dai, Elias~B Khalil, Yuyu Zhang, Bistra Dilkina, and Le~Song.
\newblock Learning combinatorial optimization algorithms over graphs.
\newblock {\em arXiv preprint arXiv:1704.01665}, 2017.

\bibitem[\protect\citeauthoryear{Defferrard \bgroup \em et al.\egroup
  }{2016}]{defferrard2016convolutional}
Micha{\"e}l Defferrard, Xavier Bresson, and Pierre Vandergheynst.
\newblock Convolutional neural networks on graphs with fast localized spectral
  filtering.
\newblock {\em arXiv preprint arXiv:1606.09375}, 2016.

\bibitem[\protect\citeauthoryear{Fey and Lenssen}{2019}]{fey2019fast}
Matthias Fey and Jan~Eric Lenssen.
\newblock Fast graph representation learning with pytorch geometric.
\newblock {\em arXiv preprint arXiv:1903.02428}, 2019.

\bibitem[\protect\citeauthoryear{Fey \bgroup \em et al.\egroup
  }{2020}]{fey2020deep}
Matthias Fey, Jan~E Lenssen, Christopher Morris, Jonathan Masci, and Nils~M
  Kriege.
\newblock Deep graph matching consensus.
\newblock {\em arXiv preprint arXiv:2001.09621}, 2020.

\bibitem[\protect\citeauthoryear{Hamilton \bgroup \em et al.\egroup
  }{2017}]{hamilton2017inductive}
William~L Hamilton, Rex Ying, and Jure Leskovec.
\newblock Inductive representation learning on large graphs.
\newblock {\em arXiv preprint arXiv:1706.02216}, 2017.

\bibitem[\protect\citeauthoryear{IBM}{2021}]{noauthor_cplex_nodate}
IBM.
\newblock {CPLEX} {Optimizer}, 2021.

\bibitem[\protect\citeauthoryear{Joshi \bgroup \em et al.\egroup
  }{2019}]{joshi2019efficient}
Chaitanya~K Joshi, Thomas Laurent, and Xavier Bresson.
\newblock An efficient graph convolutional network technique for the travelling
  salesman problem.
\newblock {\em arXiv preprint arXiv:1906.01227}, 2019.

\bibitem[\protect\citeauthoryear{Joshi \bgroup \em et al.\egroup
  }{2020}]{joshi2020learning}
Chaitanya~K Joshi, Quentin Cappart, Louis-Martin Rousseau, Thomas Laurent, and
  Xavier Bresson.
\newblock Learning tsp requires rethinking generalization.
\newblock {\em arXiv preprint arXiv:2006.07054}, 2020.

\bibitem[\protect\citeauthoryear{Kingma and Ba}{2014}]{kingma2014adam}
Diederik~P Kingma and Jimmy Ba.
\newblock Adam: A method for stochastic optimization.
\newblock {\em arXiv preprint arXiv:1412.6980}, 2014.

\bibitem[\protect\citeauthoryear{Kipf and Welling}{2016}]{kipf2016semi}
Thomas~N Kipf and Max Welling.
\newblock Semi-supervised classification with graph convolutional networks.
\newblock {\em arXiv preprint arXiv:1609.02907}, 2016.

\bibitem[\protect\citeauthoryear{Kool \bgroup \em et al.\egroup
  }{2018}]{kool2018attention}
Wouter Kool, Herke Van~Hoof, and Max Welling.
\newblock Attention, learn to solve routing problems!
\newblock {\em arXiv preprint arXiv:1803.08475}, 2018.

\bibitem[\protect\citeauthoryear{Laporte \bgroup \em et al.\egroup
  }{2019}]{laporte2019location}
Gilbert Laporte, Stefan Nickel, and Francisco Saldanha~da Gama, editors.
\newblock {\em Location Science}.
\newblock Springer, 2nd edition, 2019.

\bibitem[\protect\citeauthoryear{Li \bgroup \em et al.\egroup
  }{2015}]{li2015gated}
Yujia Li, Daniel Tarlow, Marc Brockschmidt, and Richard Zemel.
\newblock Gated graph sequence neural networks.
\newblock {\em arXiv preprint arXiv:1511.05493}, 2015.

\bibitem[\protect\citeauthoryear{Li \bgroup \em et al.\egroup
  }{2018}]{li2018combinatorial}
Zhuwen Li, Qifeng Chen, and Vladlen Koltun.
\newblock Combinatorial optimization with graph convolutional networks and
  guided tree search.
\newblock {\em arXiv preprint arXiv:1810.10659}, 2018.

\bibitem[\protect\citeauthoryear{Paszke \bgroup \em et al.\egroup
  }{2019}]{paszke2019pytorch}
Adam Paszke, Sam Gross, Francisco Massa, Adam Lerer, James Bradbury, Gregory
  Chanan, Trevor Killeen, Zeming Lin, Natalia Gimelshein, Luca Antiga, et~al.
\newblock Pytorch: An imperative style, high-performance deep learning library.
\newblock {\em arXiv preprint arXiv:1912.01703}, 2019.

\bibitem[\protect\citeauthoryear{Smith}{1999}]{smith1999neural}
Kate~A Smith.
\newblock Neural networks for combinatorial optimization: a review of more than
  a decade of research.
\newblock {\em INFORMS Journal on Computing}, 11(1):15--34, 1999.

\bibitem[\protect\citeauthoryear{Snyder and
  Shen}{2011}]{snyder2011fundamentals}
Lawrence~V Snyder and Zuo-Jun~Max Shen.
\newblock {\em Fundamentals of supply chain theory}.
\newblock Wiley Online Library, 2011.

\bibitem[\protect\citeauthoryear{Vinyals \bgroup \em et al.\egroup
  }{2015}]{vinyals2015pointer}
Oriol Vinyals, Meire Fortunato, and Navdeep Jaitly.
\newblock Pointer networks.
\newblock {\em arXiv preprint arXiv:1506.03134}, 2015.

\end{thebibliography}

\end{document}